\documentclass{article}

\usepackage{mathrsfs}\usepackage{amsmath,amssymb}\usepackage{graphicx}
\usepackage{epic,eepic}\usepackage{color}
\begin{document}
\newtheorem{example}{Example}[section]
\newtheorem{definition}{Definition}[section]
\newtheorem{theorem}{Theorem}[section]
\newtheorem{proposition}{Proposition}[section]
\newtheorem{lemma}{Lemma}[section]
\newtheorem{corollary}{Corollary}[section]
%\begin{frontmatter}

%% Title, authors and addresses

%% use the tnoteref command within \title for footnotes;
%% use the tnotetext command for the associated footnote;
%% use the fnref command within \author or \address for footnotes;
%% use the fntext command for the associated footnote;
%% use the corref command within \author for corresponding author footnotes;
%% use the cortext command for the associated footnote;
%% use the ead command for the email address,
%% and the form \ead[url] for the home page:
%%
%% \title{Title\tnoteref{label1}}
%% \tnotetext[label1]{}
%% \author{Name\corref{cor1}\fnref{label2}}
%% \ead{email address}
%% \ead[url]{home page}
%% \fntext[label2]{}
%% \cortext[cor1]{}
%% \address{Address\fnref{label3}}
%% \fntext[label3]{}

\title{Preferential Multi-Context Systems}

%% use optional labels to link authors explicitly to addresses:
%% \author[label1,label2]{<author name>}
%% \address[label1]{<address>}
%% \address[label2]{<address>}

\author{Kedian Mu$^{1}$, Kewen Wang$^{2}$, and Lian Wen$^{2}$\\
$^{1}$School of Mathematical Sciences\\Peking University,
 Beijing 100871, P. R. China\\
 $^{2}$School of Information and Communication Technology\\ Nathan Campus, Griffith University\\
170 Kessels Road, Nathan, Brisbane, Queensland 4111, Australia}

\maketitle

\begin{abstract}

Multi-context systems (MCS) presented by Brewka and Eiter can be
considered as a promising way to interlink decentralized and
heterogeneous knowledge contexts. In this paper, we propose
preferential multi-context systems (PMCS), which provide a framework
for incorporating a total preorder relation over contexts in a
multi-context system.  In a given PMCS, its contexts are divided
into several parts according to the total preorder relation over
them, moreover, only information flows from a context to ones of the
same part or less preferred parts are allowed to occur. As such, the
first $l$ preferred parts of an PMCS always fully capture the
information exchange between contexts of these parts, and then
compose another meaningful PMCS, termed the $l$-section of that
PMCS. We generalize the equilibrium semantics for an MCS to the
(maximal) $l_{\leq}$-equilibrium which represents belief states at
least acceptable for the $l$-section of an PMCS. We also investigate
inconsistency analysis in PMCS and related computational complexity
issues.

%Multi-context systems (MCS) presented by Brewka and Eiter can be
%considered as a promising way to interlink distributed and
%heterogeneous knowledge contexts. In this paper, we propose
%stratified multi-context systems (SMCS) by combining multi-context
%systems with total preorder relations on their contexts.   In a
%stratified multi-context system, contexts are grouped into several
%strata according to a total preorder relation on these contexts,
%moreover, only information flows from more preferred contexts to
%less preferred ones are allowed to occur. As such, the first $i$
%strata of an SMCS always compose a sub-SMCS, which is termed the
%$i$-section of that SMCS.  We adapt the notation of equilibrium to
%(maximal) $i$-equilibrium which represents belief states at least
%acceptable for the $i$-section of an SMCS. Then we investigate
%inconsistency analysis in SMCS. Finally, we consider computational
%complexity issues.

\end{abstract}


%%
%% Start line numbering here if you want
%%
% \linenumbers

%% main text
\section{Introduction}
\label{Introduction}

Many (if not all) real-world applications of sharing and reasoning
knowledge are characterized by heterogeneous contexts, especially
with the advent of the world wide web. Research in representing
contexts and information flow between contexts has gained much
attention recently in artificial intelligence
~\cite{DBLP:journals/ai/GiunchigliaS94,
aaaiBrewkaE07,DBLP:conf/ijcai/BrewkaRS07,DBLP:conf/kr/EiterFSW10,conf/ijcai/BrewkaEFW11,DBLP:conf/pricai/JinWW12}
as well as in applications such as requirements
engineering~\cite{ijsekeFinkelsteinKNFG92,icseNuseibehKF03,DBLP:journals/fuin/MuLJLYB09}.

Instead of finding a universal knowledge representation for all
contexts, it has been increasingly recognized that it may be
desirable to allow each context to choose a suitable representation
tool for its own to capture its knowledge precisely. For example, in
 some frameworks such as Viewpionts for eliciting and analyzing software
requirements developers often encourage stakeholders to use their
own familiar terms and notations to express their demands so as to
elicit requirements as full as
possible~\cite{ijsekeFinkelsteinKNFG92,icseNuseibehKF03}. Moreover,
the heterogeneous nature of contexts representations may allow
different monotonic or non-monotonic reasoning mechanisms to occur
together in a given system.  For example, as stated
in~\cite{aaaiBrewkaE07}, there is growing interest in combining
ontologies based on description logics with non-monotonic formalisms
in semantic web applications. However, the diversity of
representations of contexts in such cases brings some important
challenges to accessing each individual context as well as to
interlinking these contexts~\cite{conf/ijcai/BrewkaEFW11}.

Nonmonotonic multi-context systems presented by Brewka and
Eiter~\cite{aaaiBrewkaE07} can be considered as a promising way to
deal with these challenges~\cite{conf/ijcai/BrewkaEFW11}. Instead of
attempting to translate all contexts with different formalisms into
a unifying formalism, they leave the logics of contexts untouched
and interlink contexts by modeling the inter-contextual information
exchange in a uniform way. To be more precise, information flow
among contexts is articulated by so-called bridge rules in a
declarative way. Similar to logical programming rules, each bridge
rule consists of two parts, the head of the rule and the body of the
rule (possibly empty). More importantly, each bridge rule allows
access to other contexts in its body. This makes it capable of
adding information represented by its head to a context by
exchanging information with other contexts.  In semantics, several
equilibria representing acceptable belief states for multi-context
systems  are also given by Brewka and Eiter~\cite{aaaiBrewkaE07}.

Multi-context systems can be viewed as the first step towards
interlinking distributed and heterogeneous contexts effectively. The
way they operating contextual knowledge bases is only limited  to
adding information to a context when the corresponding bridge rules
are applicable~\cite{conf/ijcai/BrewkaEFW11}.   To be more
applicable to real-world applications, it is advisable to generalize
multi-context systems from some perspectives.  For example, Brewka
et al have considerably generalized multi-context systems to managed
multi-context systems (mMCS) by allowing flexible operations on
context knowledge bases~\cite{conf/ijcai/BrewkaEFW11}. Essentially,
managed multi-context systems focus on managed contexts, which are
contexts together with possible operations on them.

Combining preferences and contexts is still an interesting issue in
reasoning about contextual knowledge~\cite{iclpBrewka07}. In
particular, preferences on contexts have an important influence on
information exchange between contexts and inter-contextual knowledge
integration in many real-world applications.
%However,   preferences on contexts have an important influence on
%inter-contextual knowledge integration. In many real-world
%applications, information exchange between contexts often depends on
%preferences among contexts.
 For example, it is intuitive to revise a
less reliable knowledge base by accessing more reliable ones. But we
cannot use information deriving from  less reliable sources to
revise more reliable knowledge bases in general case.  In legal
reasoning, consequences of applying a law to a case can be rebutted
by that of applying another law with higher level when there is a
conflict, and not vise versa. In such cases, it may be advisable to
take into account preferences on contexts in  characterizing
inter-contextual information exchange in multi-context systems.

%On the other hand, taking into account the preference relation on
%contexts makes some subsets of more preferred contexts satisfying
%some given constraints more significant than others for some
%scenarios.

Moreover, taking into account the preference relation on contexts
 makes some subsets of more preferred contexts satisfying
some given constraints more significant when the whole set of
contexts does not satisfy the constraints. For example, in a
multi-party negotiation, an agreement between the most important
parties is preferred  if it is difficult to achieve an agreement
between all parties in many cases. In an incremental software
development,
 only requirements with priorities
higher than a given level are concerns of developers at a given
stage.

% However, Brewka has pointed out that it is
%useful to combine preferences and contexts in~\cite{iclpBrewka07}.

To address these issues, we combine a multi-context system with a
total preorder relation on its contexts to develop a preferential
multi-context system (PMCS) in this paper. A preferential
multi-context systems is given in the form of a sequence of sets of
contexts such that the location of a set signifies its preference
level. Without loss of generality, we assume that the smaller of the
location of a set is, the more preferred contexts in that set are.
We call each set of contexts in that sequence a stratum.  Moreover,
we assume that information flow cannot be from less preferred strata
to more preferred ones. That is, any bridge rule of a given context
does not allow any access to other strictly less preferred contexts
in its body. As such, the first several strata also compose a new
preferential multi-context system such that all the contexts
involved in it are strictly more preferred than ones out of it. We
call such a new preferential  multi-context system a section of that
system. We are interested in all sections as well as the whole
preferential
 multi-context system, and then propose $l_{\leq}$-equilibria to
represent belief sets acceptable for at least contexts in the first
$l$ strata. In particular, the maximal consistent section describes
a maximal section that has an equilibrium. Actually, it plays an
important role in inconsistency analysis in a given preferential
 multi-context system, because it can be considered as maximally
reliable part of that preferential  multi-context system. We are
more interested in finding diagnoses and inconsistency explanations
compatible with maximal consistent section instead of all ones.
Finally, we discuss computational complexity issues.

The rest of this paper is organized as follows. We give a brief
introduction to multi-context systems in Section 2. We propose
 preferential   multi-context
systems in Section 3. In section 4, we discuss inconsistency
analysis in preferential  multi-context systems.  We discuss
complexity issues in Section 5. In section 6 we compare our work
with some closely related work. Finally we conclude this paper in
Section 7.

\section{Preliminaries}

In this section, we review the details of the  definitions of
multi-context systems presented by Brewka and Eiter
~\cite{aaaiBrewkaE07} and inconsistency analysis in multi-context
systems presented in~\cite{DBLP:conf/kr/EiterFSW10}. The material is
largely taken from~\cite{aaaiBrewkaE07} and
~\cite{DBLP:conf/kr/EiterFSW10}.

The goal of multi-context systems is to  combine arbitrary monotonic
and nonmonotonic logics. Here a logic $L$ is referred to as a triple
$(KB_{L},BS_{L},ACC_{L})$, where $KB_{L}$ is the set of well-formed
knowledge bases of $L$, which characterizes the syntax of $L$;
$BS_{L}$ is the set of belief sets; and $ACC_{L}:KB_{L}\rightarrow
2^{BS_{L}}$ is a function describing the semantics of the logic by
assign to each knowledge base (a set of formulas) a set of
acceptable sets of beliefs~\cite{aaaiBrewkaE07}.

\begin{definition}{\em ~\cite{aaaiBrewkaE07}} Let $L=\{L_{1},L_{2},\cdots,L_{n}\}$ be a set of
logics. A $L_{k}$-bridge rule over $L$, $1\leq k\leq n$, is of the
form
   \begin{eqnarray*}
   (k:s)\leftarrow (r_{1}:p_{1}),\cdots,(r_{j}:p_{j}), {\bf not}\
    (r_{j+1}:p_{j+1}),\cdots, {\bf not}\ (r_{m}:p_{m})
   \end{eqnarray*}
where $1\leq r_{l}\leq n$, $p_{l}$ is an element of some belief set
of $L_{r_{l}}$, and for each $kb\in KB_{L_{k}}$, $kb\cup\{s\}\in
KB_{L_{k}}$.
\end{definition}

Similar to  logical programming rules, we call the left (resp.
right) part of $r$ the head (resp. body) of the bridge rule $r$.

\begin{definition}{\em ~\cite{aaaiBrewkaE07}} A multi-context system $M=(C_{1},C_{2},\cdots,C_{n})$
consists of a collection of contexts $C_{i}=(L_{i},kb_{i},br_{i})$,
where $L_{i}=(KB_{i},BS_{i},ACC_{i})$ is a logic, $kb_{i}\in KB_{i}$
is a knowledge base, and $br_{i}$ is a set of $L_{i}$-bridge rules
over $\{L_{1},L_{2},\cdots, L_{n}\}$.
\end{definition}

A multi-context system $M$ is finite if all knowledge bases $kb_{i}$
and sets of bridge rules $br_{i}$ are finite~\cite{aaaiBrewkaE07}.

Given a $L_{k}$-bridge rule $r$, we use $hd(r)$ to denote the head
of $r$. Further, let $cnt^{+}(r)=\{C_{r_{i}}|1\leq i\leq j\}$ and
$cnt^{-}(r)=\{C_{r_{i}}|j+1\leq i\leq m\}$. Obviously,
$cnt(r)=cnt^{+}(r)\cup cnt^{-}(r)$ is exactly the set of contexts
involved in the body of  $r$.

We use $br_{M}$ to denote the set of all bridge rules in $M$, i.e,
$br_{M}=\cup_{i=1}^{n}br_{i}$.  For any set $D\subseteq br_{M}$, we
use $heads(D)$ to denote the set of all the rules in $D$ in
unconditional form, i.e., $heads(D)=\{hd(r)\leftarrow\ | \ r\in
D\}$. Let $R$ be a set of bridge rules, we use $M[R]$ to denote the
MCS obtained from $M$ by replacing $br_{M}$ with $R$.  For a set
${\sf R}$ of sets of bridge rules, we use $\bigcup {\sf R}$ to
denote the union of all sets in ${\sf R}$.

A belief state for $M=(C_{1},C_{2},\cdots,C_{n})$ is a sequence
$S=(S_{1},S_{2},\cdots, S_{n})$ such that each $S_{i}\in BS_{i}$.  A
bridge rule $r$ is {\em applicable} in a belief state
$S=(S_{1},S_{2},\cdots, S_{n})$ iff for $1\leq i\leq j$, $p_{i}\in
S_{r_{i}}$ and for $j+1\leq l\leq m$, $p_{k}\not\in S_{r_{l}}$. We
use $app(br_{i},S)$ to denote the set of all $L_{i}$-bridge rules
that are applicable in belief state $S$.

\begin{definition}{\em ~\cite{aaaiBrewkaE07}} A belief state
$S=(S_{1},S_{2},\cdots, S_{n})$ of $M$ is an equilibrium iff, for
$1\leq i\leq n$, $S_{i}\in ACC_{i}(kb_{i}\cup\{head(r)|r\in
app(br_{i},S)\})$.
\end{definition}

Essentially, an equilibrium is a belief state which contains an
acceptable belief set for each context, given the belief sets for
other contexts~\cite{aaaiBrewkaE07}.

\begin{example} Let $M_{0}=(C_{1},C_{2},C_{3})$ be an MCS, where $L_{1}$ is a propositional logic, whilst both $L_{2}$ and
$L_{3}$ are ASP logics.  Suppose that
\begin{itemize}
\item $kb_{1}=\{a,b\}$, $br_{1}=\{(1:c)\leftarrow (2:d),(3:g)\}$;
\item $kb_{2}=\{d\leftarrow e, e\leftarrow d\}$, $br_{2}=\{ (2:p)\leftarrow
(1:c), {\bf not}\ (3:h)\}$;
\item $kb_{3}=\{f\leftarrow g,g\leftarrow\}$,  $br_{3}=\{(3:q)\leftarrow (2:p),(1: a); (3:h)\leftarrow {\bf
not}\ (1:c)\}$.
\end{itemize}

Consider $S=(\{a,b,c\},\{d,e,p\},\{f,g,q\})$. Note that all bridge
rules are applicable in $S$, except  $(3:h)\leftarrow {\bf not}\
(1:c)$.

Evidently, we can check $S$ is an equilibrium of $M_{0}$.
\end{example}

Note that it cannot be guaranteed that there exists an equilibrium
for a given multi-context system.  Inconsistency in an MCS is
referred to as the lack of an
equilibrium~\cite{DBLP:conf/kr/EiterFSW10}.  We use $M\models\bot$
to denote that $M$ is inconsistent, i.e., has no equilibrium. In
this paper, we assume that every context to be consistent if no
bridge rules apply, i.e., $M[\emptyset]\not\models\bot$.

\begin{example} Let $M_{1}=(C_{1},C_{2},C_{3})$ be an MCS, where $L_{1}$ is a propositional logic, whilst both $L_{2}$ and
$L_{3}$ are ASP logics.  Suppose that
\begin{itemize}
\item $kb_{1}=\{a,b\}$, $br_{1}=\{r_{1}=(1:c)\leftarrow (2:e)\}$;
\item $kb_{2}=\{d\leftarrow e, e\leftarrow\}$, $br_{2}=\{r_{2}=(2: p)\leftarrow
(1:c)\}$;
\item $kb_{3}=\{g\leftarrow, \bot\leftarrow q, {\bf not}\ h\}$,  $br_{3}=\{r_{3}=(3:q)\leftarrow (2:p); r_{4}=(3:h)\leftarrow
{\bf not}\ (1:a) \}$.
\end{itemize}

 Note that all bridge
rules are applicable, except  $r_{4}$. The three applicable bridge
rules in turn adds $q$ to $C_{3}$, and then activates
$\bot\leftarrow q, {\bf not}\ h$.  So,  $M_{1}$ has no equilibrium,
i.e., $M_{1}\models\bot$.
\end{example}

To analyze inconsistency, inspired by debugging approaches used in
the nonmonotonic reasoning community, T. Eiter et al have introduced
two notions of explaining inconsistency, i.e., diagnoses and
inconsistency explanations for multi-context
systems~\cite{DBLP:conf/kr/EiterFSW10}.  Roughly speaking, diagnoses
provide a consistency-based formulation for explaining
inconsistency, by finding a part of bridge rules which need to be
changed (deactivated or added in unconditional form) to restore
consistency in a multi-context system, whilst inconsistency
explanations provide an entailment-based formulation for
inconsistency, by identifying a part of bridge rules which is needed
to cause inconsistency~\cite{DBLP:conf/kr/EiterFSW10}.

\begin{definition}{\em ~\cite{DBLP:conf/kr/EiterFSW10}}
Given an MCS $M$, a diagnosis of $M$ is a pair $(D_{1},D_{2})$,
$D_{1},D_{2}\subseteq br_{M}$, s.t. $M[br_{M}\setminus D_{1}\cup
heads(D_{2})]\not\models\bot$. $D^{\pm}(M)$ is the set of all such
diagnosis.
\end{definition}

Essentially, a diagnosis exactly captures a pair of sets of bridge
rules such that inconsistency will disappear if we deactivate the
rules in the first set, and add the rules in the second set in
unconditional form~\cite{DBLP:conf/kr/EiterFSW10}.

\begin{definition}{\em ~\cite{DBLP:conf/kr/EiterFSW10}}
$D^{\pm}_{m}(M)$ is the set of all pointwise subset-minimal
diagnoses of an MCS $M$, where the pointwise subset relation
$(D_{1},D_{2})\subseteq (D'_{1},D'_{2})$ holds iff $D_{1}\subseteq
D'_{1}$ and $D_{2}\subseteq D'_{2}$.
\end{definition}

\begin{example} Consider $M_{1}$ again. Then
$$D^{\pm}_{m}(M_{1})=\{(\{r_{1}\},\emptyset),(\{r_{2}\},\emptyset),(\{r_{3}\},\emptyset),(\emptyset,\{r_{4}\})\}.$$
This means we need only to deactivate one of  $r_{1}$, $r_{2}$, and
$r_{3}$, or to add $r_{4}$ unconditionally, in order to restore
consistency for $M_{1}$.
\end{example}

\begin{definition}{\em ~\cite{DBLP:conf/kr/EiterFSW10}} Given an MCS
$M$, an inconsistency explanation of $M$ is a pair $(E_{1},E_{2})$
of sets $E_{1},E_{2}\in br_{M}$ of bridge rules s.t. for all
$(R_{1},R_{2})$ where $E_{1}\subseteq R_{1}\subseteq br_{M}$ and
$R_{2}\subseteq br_{M}\setminus E_{2}$, it holds that $M[R_{1}\cup
heads(R_{2})]\models\bot$. By $E^{\pm}(M)$ we denote the set of all
inconsistency explanations of $M$, and by $E^{\pm}_{m}(M)$ the set
of all pointwise subset-minimal ones.
\end{definition}

Essentially, an inconsistency explanation captures a pair of sets of
bridge rules such that the rules in the first set cause an
inconsistency relevant to the MCS, and this inconsistency cannot be
resolved by adding bridge rules unconditionally, unless we use at
least one bridge rule in the second
set~\cite{DBLP:conf/kr/EiterFSW10}.

\begin{example} Consider $M_{1}$ again. Then
$$E^{\pm}_{m}(M_{1})=\{(\{r_{1},r_{2},r_{3}\},\{r_{4}\})\}.$$
This means that the inconsistency in $M_{1}$ is caused by $r_{1}$,
$r_{2}$, and $r_{3}$ together, moreover, it can be resolved by
adding $r_{4}$ unconditionally.
\end{example}

Note that both addition and removal of knowledge can prevent
inconsistency in nonmonotonic reasoning. So, a diagnosis consists of
two sets of bridge rules including the set of bridge rules to be
removed and that to be added unconditionally.  As pointed out in
~\cite{DBLP:conf/kr/EiterFSW10}, for scenarios  where removal of
bridge rules is preferred to unconditional addition of rules, we may
focus on diagnoses of the form $(D_{1},\emptyset)$ only.

\begin{definition}{\em ~\cite{DBLP:conf/kr/EiterFSW10}} Given an MCS $M$, an $s$-diagnosis of $M$ is a
set $D\subseteq br_{M}$ s.t. $M[br_{M}\setminus D]\not\models\bot$.
The set of all $s$-diagnoses (resp., $\subseteq$-minimal
$s$-diagnoses) is $D^{-}(M)$ (resp., $D^{-}_{m}(M)$).
\end{definition}

Similarly, we need only focus on inconsistency  explanations in form
of $(E_{1},br_{M})$ if adding rules unconditionally is less
preferred.

\begin{definition}{\em ~\cite{DBLP:conf/kr/EiterFSW10}} Given an MCS $M$, an $s$-inconsistency explanation of $M$ is a
set $E\subseteq br_{M}$ s.t. each $R$ where $E\subseteq R\subseteq
br_{M}$, satisfies  $M[R]\models\bot$. The set of all
$s$-inconsistency explanations (resp., $\subseteq$-minimal
$s$-inconsistency explanations) is $E^{+}(M)$ (resp.,
$E^{+}_{m}(M)$).
\end{definition}

\begin{example} Consider $M_{1}$ again. Then
$$D^{-}_{m}(M_{1})=\{\{r_{1}\},\{r_{2}\},\{r_{3}\}\},\ \ E^{+}_{m}(M_{1})=\{\{r_{1},r_{2},r_{3}\}\}.$$
\end{example}

More interestingly, Eiter  et al have obtained the following duality
relation between diagnoses and inconsistency explanations:

\begin{theorem}~\label{em+=dm-} {\em
~\cite{DBLP:conf/kr/EiterFSW10}} Given an inconsistent MCS $M$,
$$\bigcup D^{\pm}_{m}(M)=\bigcup E^{\pm}_{m}(M),$$ and $$\bigcup
D^{-}_{m}(M)=\bigcup E^{+}_{m}(M).$$
\end{theorem}

This duality theorem shows that the unions of all minimal diagnoses
and all inconsistency explanations coincide, i.e., diagnoses and
inconsistency explanations represent dual aspects of inconsistency
in an MCS~\cite{DBLP:conf/kr/EiterFSW10}.

\section{Preferential Multi-context Systems}

In this section we formally introduce a class of MCSs that allows us
to consider preference information on contexts, called
\emph{preferential multi-context systems}, or simply PMCSs. As
explained in the introduction, the motivation for such MCSs is that
in many practical applications, it is often the case that some
context has higher priority over another context. For example, the
ontology SNOWMED CT (a context) will have higher priority over
Wikipedia (another context) for medical doctors. In the setting of
MCSs, an PMCS $P$ is a pair $(M,\leq_s)$ such that the following
conditions are satisfied:
\begin{enumerate}
\item [(1)] $M$ is an MCS that has a splitting $M=\cup_{i=1}^{m}M_{i}$.
\item [(2)] $\leq_s$ is a total preorder\footnote{ A binary relation $\leq$ on some set $A$
is a total preorder relation if it is reflexive, transitive, and
total, i.e., for all $a,b,c\in A$, we have that:
\begin{enumerate}
\item [(1)] $a\leq a$ (reflexivity),
\item [(2)]if $a\leq b$ and $b\leq c$, then $a\leq c$
(transitivity),
\item [(3)] $a\leq b$ or $b\leq a$ (totality).
\end{enumerate}
} on the set $\{M_1, \ldots, M_m\}$.
\end{enumerate}

Recall that $M=\cup_{i=1}^{m}M_{i}$ is a splitting for $M$ if
$M_i\neq\emptyset$ for all $i$ and $M_{i}\cap M_{j}=\emptyset$ for
all $i\neq j$.
%$\leq_s$ is a total preorder on the set $\{M_1,
%\ldots, M_m\}$.

Informally, $M_i\leq_s M_j$ means that a context $C$ in $M_i$ is always preferred to a context $C'$ in $M_j$.
%In this case, we write $C\leq_s C'$, which is intended to express that bridge rules in $C$ have higher priority over bridge rules in $C'$. Thus, $\leq_s$ induces a preference relation on the set $br_M$ of bridge rules in the MCS $M$: $r\leq_s r'$ if $r\in C$, $r'\in C'$ and $C\leq_s C'$.
 We assume that the smaller a subscript $i$ is , the
more preferred
 $M_{i}$ is. Then we use $\langle M_1, \ldots, M_m\rangle$ instead
 of $\{M_1, \ldots, M_m\}$ from now on.

In an PMCS, preference information controls the information flow
from one context to another context. Specifically, a context can be
impacted only by  more or equally preferred ones. This notion is
formally defined as follows.
\begin{definition} Let $\leq_{s}$ be  a total preorder relation on the
set of contexts  $M=\langle M_{1}, M_{2},\cdots, M_{m}\rangle$.
\begin{enumerate}
\item [{\em (1)}] The set $br_{l}$ of bridge rules of $C_{l}\in M_{i}$ is {\em
compatible} with the preorder relation $\leq_{s}$ on $M$ if for all
$r\in br_{l}$, $cnt(r)\cap M_{j}=\emptyset$  for all $j>i$.
\item [{\em (2)}] The set $br_{M}$ of bridge rules of $M$ is {\em compatible} with the
preorder relation $\leq_{s}$ on $M$ if $br_{i}$ is {\em compatible}
with $\leq_{s}$ for all $1\leq i\leq n$.
\end{enumerate}
\end{definition}

Essentially, the compatibility of $br_{l}$ with $\leq_{s}$ implies
that only information exchange between $C_{l}$ with some $C_{k}$s
satisfying $C_{k}\leq C_{l}$ for each $k$ may activate possible
change of $kb_{l}$ in  $C_{l}$.

Given an MCS $M$ and a total preorder relation $\leq_{s}$ on
contexts in $M$,  we say that $M$ is {\em compatible with}
$\leq_{s}$ iff $br_{M}$  is compatible with $\leq_{s}$.

\begin{definition}[Preferential multi-context system] A preferential
multi-context system (PMCS) is a pair $(M,\leq_{s})$, where $M$ is
an MCS, and $\leq_{s}$ is a total preorder relation on contexts in
$M$ such that $M$ is compatible with $\leq_{s}$.
\end{definition}

An PMCS  $(M,\leq_{s})$ is represented in the form of a sequence
$\langle M_{1},M_{2},\cdots, M_{m}\rangle$ such that for
$C_{i},C_{j}\in M$, $C_{i}\leq_{s} C_{j}$ iff for some
$t_{i},t_{j}$: $C_{i}\in M_{t_{i}}$, $C_{j}\in M_{t_{j}}$ and
$t_{i}\leq t_{j}$.  In particular, we may consider an MCS $M$ as a
special PMCS $(M,\emptyset)$, which contains only one stratum, i.e.,
$\langle M\rangle$.

Essentially, preferential multi-context systems take into account
the impact of preference relation over contexts on inter-contextual
information exchange.  Only information flow from a context to
equally or less preferred ones are allowed to occur  in preferential
multi-context systems.

Let $(M,\leq_{s})=\langle M_{1},M_{2},\cdots, M_{m}\rangle$ be an
PMCS. Then the $i$-cut of $(M,\leq_{s})$ for each $1\leq i\leq m$,
denoted $M(i)$, is defined as $M(k)=\bigcup\limits_{k=1}^{i}M_{k}$.
Correspondingly, we call $M_{1\rightarrow i}=\langle
M_{1},M_{2},\cdots, M_{i}\rangle$ the $i$-section of $(M,\leq_{s})$.
Note that the compatibility of $M$ and $\leq_{s}$ ensures that each
$i$-section of $(M,\leq_{s})$ is also an PMCS. Correspondingly, each
$i$-cut of $(M,\leq_{s})$ is an MCS.  Informally speaking, given an
PMCS, the $i$-section is exactly the PMCS consisting of the first
$i$ strata in $(M,\leq_{s})$, in which all the contexts are
preferred to ones in $M_{i+1}$ for each $1\leq i\leq m-1$.  This
implies that the $i$-section of an PMCS exactly capture the
inter-contextual information exchange between contexts preferred to
ones in $M_{i+1}$.

A belief state for $\langle M_{1},M_{2},\cdots, M_{m}\rangle$ is a
sequence $\mathcal{S}=\langle \mathcal{S}_{1},
\mathcal{S}_{2},\cdots, \mathcal{S}_{m}\rangle$ such that
$\mathcal{S}_{1}\uplus\cdots\uplus\mathcal{S}_{i}$ is a belief state
of $M(i)$ for all $1\leq i\leq m$, where $\uplus$ is a concatenation
operator. In particular, we use $\uplus\mathcal{S}$ to denote
$\mathcal{S}_{1}\uplus\cdots\uplus\mathcal{S}_{m}$.

\begin{definition} A belief state $\mathcal{S}=\langle \mathcal{S}_{1},
\mathcal{S}_{2},\cdots, \mathcal{S}_{m}\rangle$  of $(M,\leq_{s})$
is an equilibrium of $(M,\leq_{s})$ iff  $\uplus\mathcal{S}$ is an
equilibrium of $M$.
\end{definition}

\begin{example} Consider an PMCS  $(M_{2},\leq_{s})=\langle
(C_{1},C_{2}),(C_{3}),(C_{4},C_{5})\rangle$,  where $L_{1}$ and
$L_{2}$ are propositional logics, and others are ASP logics. Suppose
that
\begin{itemize}
\item $kb_{1}=\{a\}$, $br_{1}=\{r_{11}=(1:c)\leftarrow (2:b)\}$;
\item $kb_{2}=\{b\}$, $br_{2}=\{r_{21}=(2:d)\leftarrow (1:a)\}$;
\item $kb_{3}=\{e\leftarrow f\}$, $br_{3}=\{r_{31}=(3:f)\leftarrow (1:c),r_{32}=(3:d)\leftarrow\ {\bf not}\ (2: b)\}$;
\item $kb_{4}=\{g\leftarrow h,h\leftarrow g\}$, $br_{4}=\{r_{41}=(4:h)\leftarrow (1:a),
(3:f)\}$;
\item $kb_{5}=\{p\leftarrow\}$, $br_{5}=\{r_{51}=(5:q)\leftarrow (2:d), (3:f), {\bf not}\ (4:\neg
h)\}$.
\end{itemize}

Consider
$\mathcal{S}=\langle(\{a,c\},\{b,d\}),(\{e,f\}),(\{h,g\},\{p,q\})\rangle$.
Then all bridge rules are applicable in $\mathcal{S}$ except
$r_{32}$. Moreover, it is easy to check that $\mathcal{S}$ is an
equilibrium of $(M_{2},\leq_{s})$.
\end{example}

\begin{figure}[thb]%[h]
\setlength{\unitlength}{0.15cm}
  \begin{picture}(60,30)
\thinlines \put(15,1){\dashbox(55,8)} \put(52,6){{\scriptsize
stratum 3: $(C_{4},C_{5})$}}
\put(24,5){\circle{4}}\put(23,4.5){{\scriptsize $C_{4}$}}
\put(46,5){\circle{4}}\put(45,4.5){{\scriptsize $C_{5}$}}
\put(26,5){\vector(1,0){18}}

\put(15,12){\dashbox(55,8)}\put(52,17){{\scriptsize stratum 2:
$(C_{3})$}} \put(35,16){\circle{4}}\put(34,15.5){{\scriptsize
$C_{3}$}}

\put(15,23){\dashbox(55,8)}\put(52,28){{\scriptsize stratum 1:
$(C_{1},C_{2})$}} \put(24,27){\circle{4}}\put(23,26.5){{\scriptsize
$C_{1}$}} \put(46,27){\circle{4}}\put(45,26.5){{\scriptsize
$C_{2}$}} \put(26,27){\vector(1,0){18}}
\put(44,27){\vector(-1,0){18}}

\thicklines
 \put(24,25){\vector(0,-1){18}}
\put(46,25){\vector(0,-1){18}}
\put(33,16){\vector(-1,-1){9}}\put(37,16){\vector(1,-1){9}}\put(24,25){\vector(1,-1){9}}\put(46,25){\vector(-1,-1){9}}

  \end{picture}
\caption{Information flow in
$(M_{2},\leq_{s})$}\label{fig:information flow}
\end{figure}
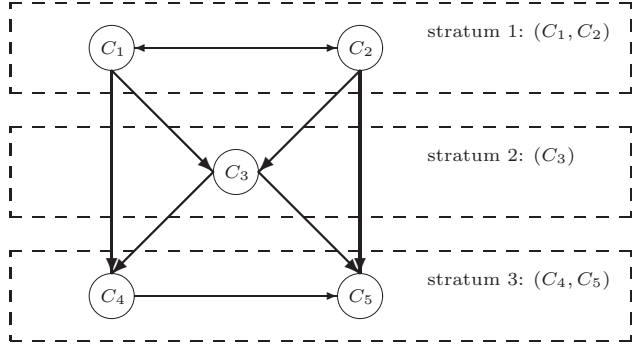

On the other hand, we can use a directed graph $G=(V,E)$ to
illustrate the information flow in a (preferential) multi-context
system $M=(C_{1},\cdots,C_{n})$, where $V=\{C_{1},\cdots, C_{n}\}$,
and $\langle C_{i},C_{j} \rangle\in E$ if $\exists r\in br_{j}$ s.t.
$C_{i}\in cnt(r)$. For example,  the information flow in
$(M_{2},\leq_{s})$ is illustrated in Figure~\ref{fig:information
flow}. Note that in such an information flow graph, there is at most
one  edge between any two contexts belonging to  different strata,
moreover, such an edge must be from a preferred context to another
context.

As mentioned in~\cite{conf/ijcai/BrewkaEFW11}, inter-contextual
information exchange among decentralized and heterogeneous contexts
can cause an MCS to be inconsistent. Moreover, inconsistency in an
MCS renders the system useless. However, in the case of preferential
multi-context systems, inconsistency may not be considered as a
totally undesirable. Allowing for preferences on contexts, we are
more interested in some consistent sections of an inconsistent PMCS,
which are significant in some applications. To address this issue,
we generalize the notion of equilibrium to an $l_{\leq}$-equilibrium
for an PMCS as follows.

\begin{definition}[$l_{\leq}$-equilibrium] Given an PMCS $(M,\leq_{s})=\langle M_{1},M_{2},\cdots, M_{m}\rangle$ and a number $l\in\{1,2,\cdots,m\}$. A belief state $\mathcal{S}=\langle \mathcal{S}_{1},
\mathcal{S}_{2}\cdots, \mathcal{S}_{m}\rangle$ of $(M,\leq_{s})$ is
an $l_{\leq}$-equilibrium of $(M,\leq_{s})$ iff $\langle
\mathcal{S}_{1}, \mathcal{S}_{2}\cdots, \mathcal{S}_{l}\rangle$ is
an equilibrium of the $l$-section $M_{1\rightarrow l}$ of
$(M,\leq_{s})$.
\end{definition}

Roughly speaking, an $l_{\leq}$-equilibrium of a preferential
 multi-context system represents belief sets acceptable for at least
all the contexts in the first $l$ strata of $(M,\leq_{s})$, given
the belief sets for other contexts.  Note that an
$l_{\leq}$-equilibrium of $(M,\leq_{s})$  must be an
$k_{\leq}$-equilibrium for all $k\leq l$. In particular,
 an equilibrium of $(M,\leq_{s})$ is an $l_{\leq}$-equilibrium
of $(M,\leq_{s})$ for all $1\leq l\leq m$. But it does not hold vice
versa.

\begin{definition}[$l_{<}$-equilibrium]  Given an PMCS $(M,\leq_{s})=\langle M_{1},M_{2},\cdots, M_{m}\rangle$ and a number $l\in\{1,2,\cdots,m\}$.  A belief state $\mathcal{S}=\langle \mathcal{S}_{1},
\mathcal{S}_{2}\cdots, \mathcal{S}_{m}\rangle$ of $(M,\leq_{s})$ is
called an  $l_{<}$-equilibrium of $(M,\leq_{s})$ iff
\begin{itemize}
  \item $\mathcal{S}$
is an $l_{\leq}$-equilibrium of $(M,\leq_{s})$,
\item but  $\mathcal{S}$ is not an
$(l+1)_{\leq}$-equilibrium of $(M,\leq_{s})$ if $l+1\leq m$.
\end{itemize}
\end{definition}

Essentially, an $l_{<}$-equilibrium of a preferential multi-context
system $(M,\leq_{s})$ represents belief sets acceptable for all the
contexts in the first $l$ strata of $(M,\leq_{s})$, but not for at
least one context in the $(l+1)$-stratum if $l<m$, given the belief
sets for other contexts. Evidently, any equilibrium of
$(M,\leq_{s})$ is an $m_{<}$-equilibrium according to this
definition.

\begin{definition}[Maximal $l_{<}$-equilibrium] Given an PMCS $(M,\leq_{s})=\langle M_{1},M_{2},\cdots, M_{m}\rangle$ and a number $l\in\{1,2,\cdots,m\}$.   A belief state $\mathcal{S}=\langle \mathcal{S}_{1},
\mathcal{S}_{2}\cdots, \mathcal{S}_{m}\rangle$ of $(M,\leq_{s})$ is
called a maximal  $l_{<}$-equilibrium of $(M,\leq_{s})$ iff
    \begin{itemize}
     \item  $\mathcal{S}$ is an $l_{<}$-equilibrium of $(M,\leq_{s})$,
     \item  For any  $i_{<}$-equilibrium $\mathcal{S}'$ of $(M,\leq_{s})$, $i\leq l$.
  \end{itemize}
\end{definition}

Actually, a maximal $l_{<}$-equilibrium of a preferential
 multi-context system is indeed an equilibrium of that system if that
system is consistent, otherwise, it represents belief sets
acceptable for contexts in a section which cannot keep consistent if
we add the next stratum to it.

\begin{example} Consider an PMCS $$(M_{3},\leq_{s})=\langle
(C_{1},C_{2}),(C_{3},C_{4}),(C_{5}),(C_{6})\rangle,$$ where $L_{1}$,
$L_{2}$, and $L_{6}$ are propositional logics, and others are ASP
logics. Suppose that
\begin{itemize}
\item $kb_{1}=\{a\}$, $br_{1}=\{r_{11}=(1:c)\leftarrow (2:b)\}$;
\item $kb_{2}=\{b\}$, $br_{2}=\{r_{21}=(2:d)\leftarrow (1:a)\}$;
\item $kb_{3}=\{e\leftarrow f\}$, $br_{3}=\{r_{31}=(3:f)\leftarrow (1:c)\}$;
\item $kb_{4}=\{g\leftarrow h,h\leftarrow g\}$, $br_{4}=\{r_{41}=(4:h)\leftarrow
(2:d),{\bf not}\ (1:b)\}$;
\item $kb_{5}=\{m\leftarrow,\bot\leftarrow q, {\bf not}\  p\}$,  $br_{5}=\{r_{51}=(5:q)\leftarrow
(3:f);
r_{52}=(5:p)\leftarrow {\bf not }\ (2:b)$;\}
\item $kb_{6}=\{\neg r\}$, $br_{6}=\{r_{61}=(6:r)\leftarrow (4:h)\}$.
\end{itemize}

Evidently, all bridge rules are applicable except $r_{52}$.
Moreover, applying $r_{11}$, $r_{31}$, and  $r_{51}$ in turn adds
$q$ to $C_{5}$, and then activates $\bot\leftarrow q, {\bf not}\ p$.
On the other hand,  applying $r_{21}$, $r_{41}$, and $r_{61}$ in
turn adds $r$ to $C_{6}$, and then results in both $r$ and $\neg r$
occurring in $C_{6}$. So,  $(M_{3},\leq_{s})$ has no equilibrium,
i.e., $M_{3}\models\bot$. Moreover, it also implies that its
$3$-section also has no equilibrium, i.e.,
$\langle(C_{1},C_{2}),(C_{3},C_{4}),(C_{5})\rangle\models\bot$.

However, both the $1$-section and $2$-section of $(M_{3},\leq_{s})$
are consistent. Obviously, we can check
\begin{itemize}
\item ${\mathcal{S}}_{0}=\langle
(\{a,c\},\{b,d\}),(\{e\},\{h\}),(\{m,q\}),(\{r\})\rangle$ is an
$1_{\leq}$-equilibrium, but not an  $2_{\leq}$-equilibrium; So, it
is an $1_{<}$-equilibrium.
\item ${\mathcal{S}}_{1}=\langle
(\{a,c\},\{b,d\}),(\{e,f\},\{g,h\}),(\{m,q\}),(\{r\})\rangle$ is an
$2_{<}$-equilibrium;
\item  ${\mathcal{S}}_{1}$ is a maximal $2_{<}$-equilibrium of
$(M_{3},\leq_{s})$.
\end{itemize}

\end{example}

An occurrence of inconsistency in a multi-context system makes that
system useless. However, considering preferences in preferential
multi-context systems makes things better.  The section
corresponding to a maximal $l_{<}$-equilibrium may be interesting
and useful in the presence  of inconsistency, because it fully
captures the meaningful information exchange among contexts involved
in this section.

\section{Inconsistency Analysis}

Now an interesting question arises: how to measure the degree of
inconsistency for an PMCS?  Note that the value $l+1$ points out the
stratum where we first meet inconsistency if a given inconsistent
PMCS $(M,\leq_{s})$ has a maximal $l_{<}$-equilibrium.  In
particular, if we abuse the notation and say that $(M,\leq_{s})$ has
a maximal $0_{<}$-equilibrium if it has no maximal
$l_{<}$-equilibrium for any given $1\leq l\leq m$.  Then $l+1$ is
exactly the inconsistency rank for stratified knowledge bases
presented in~\cite{BenferhatKBW01,BenferhatKBW04} in essence. To
bear this in mind, we  present the following inconsistency measure.

\begin{definition}   Given an PMCS $(M,\leq_{s})=\langle M_{1},M_{2},\cdots,
M_{m}\rangle$.  The degree of inconsistency of  $(M,\leq_{s})$,
denoted ${\rm DI}((M,\leq_{s}))$,  is defined as
\begin{eqnarray*}
{\rm DI}((M,\leq_{s}))=1-\frac{l}{m},
\end{eqnarray*}
if $(M,\leq_{s})$ has the maximal $l_{<}$-equilibrium, where $0\leq
l\leq m$.
\end{definition}

%%%%

Actually, the degree of inconsistency ${\rm DI}((M,\leq_{s}))$ of
$(M,\leq_{s})$ is a slight adaptation of the inconsistency rank
 such that
\begin{itemize}
\item $0\leq {\rm DI}((M,\leq_{s}))\leq 1$;
\item ${\rm DI}((M,\leq_{s}))=0$ iff $(M,\leq_{s})$ is consistent;
\item ${\rm DI}((M,\leq_{s}))=1$ iff $M_{1}\models\bot$.
\end{itemize}
Note that the first two properties are called {\em Normalization}
and {\em Consistency}, respectively~\cite{Hunter2010}.  The third
property says that an PMCS has the upper bound $1$ iff there is no
consistent section.

\begin{example}
Consider $(M_{3},\leq_{s})$ again. Note that
 $${\rm DI}((M_{3},\leq_{s}))=1-\frac{2}{4}=\frac{1}{2},$$  because it has an maximal
$2_{<}$-equilibrium as illustrated above.
\end{example}

The measure ${\rm DI}((M,\leq_{s}))$ allows us to have a sketchy
picture on the inconsistency in $(M,\leq_{s})$.  In many
applications, we need to find more information about the
inconsistency. For example, we need to know which contexts and
bridge rules of a given PMCS are involved in the inconsistency in
order to restore consistency of the PMCS.

Note that any two contexts are considered equally preferred in
inconsistency handling in the case of multi-context systems.
However, preferences over contexts play an important role in dealing
with inconsistency among these contexts, especially in making some
tradeoff decisions on resolving inconsistency when we take into
account preferences. Generally, the more preferred contexts are
considered more reliable when an inconsistency occurs in a
preferential   multi-context system, moreover, remaining unchanged
is preferred to any action of revision for  such contexts. For
example, in requirements engineering, when two requirements with
different priority levels contradict each other, a less preferred
requirement will be revised to accommodate itself to another one in
most cases.

Given an PMCS, each section actually splits the whole set of
contexts into two parts, i.e., itself and a set of other strictly
less preferred contexts. Moreover, each consistent section fully
captures information exchange among contexts which are strictly
preferred to ones not included in that section.  Generally, such a
section may be considered as one of plausible parts of that PMCS.
Allowing for this,  we are more interested in a section that
contains more preferred strata as much as possible. Moreover, any
changes of bridge rules for restoring consistency should not affect
information exchange among contexts in such a section. In this
sense, identifying a consistent section with the maximal number of
strata is central to inconsistency analysis in a preferential
multi-context system.

\begin{definition} [Maximal consistent section] Given an PMCS
$(M,\leq_{s})=\langle M_{1},M_{2},\cdots, M_{m}\rangle$, the
$i$-section $M_{1\rightarrow i}$ of $(M,\leq_{s})$, is called a
maximal consistent section of $(M,\leq_{s})$, if
  \begin{itemize}
   \item $M_{1\rightarrow i}\not\models\bot$;
   \item $M_{1\rightarrow k}\models\bot$ for all $k>i$.
  \end{itemize}
\end{definition}

Informally speaking, the maximal consistent section of an PMCS can
be considered as a reliable part of that PMCS.  We use
$M_{1\rightarrow {\rm k_{mc}}}$ to denote the maximal consistent
section of $(M,\leq_{s})$. Evidently, given an inconsistent PMCS
$(M,\leq_{s})=\langle M_{1},M_{2},\cdots, M_{m}\rangle$, a maximal
$l_{<}$-equilibrium of $(M,\leq_{s})$ is exactly an equilibrium of
the $l$-section $M_{1\rightarrow l}$, because less preferred
contexts cannot bring new information to more preferred contexts in
an PMCS. This implies that finding the maximal consistent section
may be not harder than finding maximal $l_{<}$-equilibrium.

\begin{example} Consider $(M_{3},\leq_{s})$ again. The $2$-section
$\langle(C_{1},C_{2}),(C_{3},C_{4})\rangle$ is its maximal
consistent section.
\end{example}

As mentioned above, Eiter et al have proposed diagnoses and
inconsistency explanations for a multi-context system. We use the
following example to demonstrate what will happen when we apply
these to a preferential multi-context system.

\begin{example} Consider $(M_{3},\leq_{s})$ again. Note that all of
the following sets of rules are $\subseteq$-minimal $s$-diagnoses of
$(M_{3},\leq_{s})$:
\begin{itemize}
\item $D_{1}=\{r_{51},r_{61}\}$, $D_{2}=\{r_{51},r_{41}\}$,$D_{3}=\{r_{51},r_{21}\}$;
\item $D_{4}=\{r_{31},r_{61}\}$, $D_{5}=\{r_{31},r_{41}\}$, $D_{6}=\{r_{31},r_{21}\}$;
\item $D_{7}=\{r_{11},r_{61}\}$, $D_{8}=\{r_{11},r_{41}\}$,
$D_{9}=\{r_{11},r_{21}\}$.
\end{itemize}
Note that all of the $\subseteq$-minimal $s$-diagnoses contains one
bridge rule of maximal consistent section except $D_{1}$.  That is,
according to $D_{i}$ for all $i\geq 2$, we need to deactivate some
information exchange in maximal consistent section to restore
consistency in $(M_{3},\leq_{s})$. In contrast, $D_{1}$ leaves
information exchange in maximal consistent section unchanged.
Allowing for preferences relation over contexts, $D_{1}$ is more
significant for inconsistency handling in  $(M_{3},\leq_{s})$.
\end{example}

The example above illustrates that diagnoses not involving maximal
consistent section in inconsistency are more preferred.
  Allowing
for the duality relation between diagnoses and explanations, we have
the same opinion on inconsistency explanations.  However, the
compatibility to more preferred knowledge is considered as one of
useful strategies in preferential knowledge revision and
integration~\cite{BenferhatKBW01,BenferhatKBW04}.  Next we adapt
diagnoses and inconsistency explanations to accommodate maximal
consistent section, respectively.

\begin{definition} Given an PMCS $(M,\leq_{s})$, a diagnosis $(D_{1},D_{2})$ of
$M$ is compatible to the maximal consistent section of
$(M,\leq_{s})$  if  $(D_{1}\cup D_{2})\cap br_{M({\rm
k_{mc}})}=\emptyset$.
\end{definition}

Note that if we focus on the maximal consistent section of a
preferential multi-context system, then the set $br_{M}\setminus
br_{M({\rm k_{mc}})}$ of bridge rules of all contexts out of the
section
 exactly composes a diagnosis  $(br_{M}\setminus
br_{M({\rm k_{mc}})},\emptyset)$ of inconsistency for that system,
because $M[br_{M({\rm k_{mc}})}]\not\models\bot$.  This guarantees
that there exists at least one diagnosis compatible with the maximal
consistent section.

\begin{example}
Consider $(M_{3},\leq_{s})$ again. All of
$(\{r_{51},r_{61},r_{52}\},\emptyset)$,
$(\{r_{51},r_{61}\},\emptyset)$ and
    $(\{r_{61}\},\{r_{52}\})$ are diagnoses compatible to the maximal consistent
    section.
\end{example}

Furthermore, we consider  minimal diagnoses compatible with the
maximal consistent section of a given PMCS.

\begin{definition} [$c$-diagnosis] Given an PMCS
$(M,\leq_{s})$, an $s$-diagnosis $D$ of $M$, is called an
$c$-diagnosis of $(M,\leq_{s})$, if $D\in D^{-}_{m}(M)$ and $D\cap
br_{M({\rm k_{mc}})}=\emptyset$. The set of all $c$-diagnosis of
$(M,\leq_{s})$ is $D_{c}^{-}((M,\leq_{s}))$.
\end{definition}

Essentially, an $c$-diagnosis $D$ of $(M,\leq_{s})$ is an
$\subseteq$-minimal $s$-diagnosis that is compatible with the
maximal consistent section of $(M,\leq_{s})$, i.e.,  none of bridge
rules of the maximal consistent section of $(M,\leq_{s})$ is
involved in $D$ .

\begin{example}
Consider $(M_{3},\leq_{s})$ again. Then $D_{1}=\{r_{51},r_{61}\}$ is
a unique $c$-diagnosis compatible to the maximal consistent section,
i.e., $D_{c}^{-}((M_{3},\leq_{s}))=\{D_{1}\}$.
\end{example}

Note that for all $D\in D_{c}^{-}((M,\leq_{s}))$, $D\in
D_{m}^{-}((M,\leq_{s}))$ and $D\cap br_{M({\rm k_{mc}})}=\emptyset$.
So, $\bigcup D_{c}^{-}((M,\leq_{s}))\subseteq\bigcup
D_{m}^{-}((M,\leq_{s}))\setminus br_{M({\rm k_{mc}})}$, but not vice
versa.

\begin{definition} [$c$-inconsistency explanation]  Given an PMCS
$(M,\leq_{s})$, an $c$-inconsistency explanation  $E$ of
$(M,\leq_{s})$, is a set $E\subseteq br_{M}$ s.t. each $E\subseteq
R\subseteq br_{M}\setminus br_{M({\rm k_{mc}})}$, satisfies
$M[br_{M({\rm k_{mc}})}\cup R]\models\bot$.
 The set of all $\subseteq$-minimal $c$-inconsistency explanations of
$(M,\leq_{s})$ is $E_{c}^{+}((M,\leq_{s}))$.
\end{definition}

Essentially, an $c$-inconsistency explanation focuses on the set of
other bridges rules need to cause an inconsistency given a set of
bridge rules of the maximal consistent section. Both
$c$-inconsistency explanations and $c$-diagnoses capture the
inconsistency under an assumption that every bridge rule of the
maximal consistent section should not  be revised or modified to
restore consistency.

\begin{example}
Consider $(M_{3},\leq_{s})$ again. Then both $E_{1}=\{r_{51}\}$ and
$E_{2}=\{r_{61}\}$ are $\subseteq$-minimal $c$-inconsistency
explanations compatible to the maximal consistent section, moreover,
$E_{c}^{+}((M_{3},\leq_{s}))=\{E_{1},E_{2}\}$.
\end{example}

More interestingly, we have the following  weak duality relation
between $c$-diagnoses and $c$-inconsistency explanations.

\begin{proposition} Given an inconsistent PMCS  $(M,\leq_{s})$, then  $$\bigcup
E^{+}_{c}((M,\leq_{s}))=\bigcup D^{-}_{c}((M,\leq_{s})).$$
\end{proposition}

\paragraph{Proof}  This is a direct consequence of Theorem
~\ref{em+=dm-} in essence. The main part of this proof is the same
as that of Theorem ~\ref{em+=dm-} provided
in~\cite{DBLP:conf/kr/EiterFSW10}.

Let $(M,\leq_{s})$ be an PMCS and $M_{1\rightarrow {\rm k_{mc}}}$
its maximal consistent section. The complement of $R$ w.r.t.
$br_{M}$ is denoted as $\overline{R}=br_{M}\setminus R$.

We first prove that $\bigcup E^{+}_{c}((M,\leq_{s}))\supseteq
\bigcup D^{-}_{c}((M,\leq_{s}))$ holds.   Let $D\in
D^{-}_{c}((M,\leq_{s}))$, then $br_{M({\rm k_{mc}})}\subseteq
\overline{D}$. We show that there exists $E\in E^{+}_{c}$ with $x\in
E$, for $x\in D$.

Consider $\widetilde{E}=\overline{D\setminus\{x\}}$, then
$br_{M({\rm k_{mc}})}\subset \widetilde{E}$ and
$M[\widetilde{E}]\models \bot$. Let $E=\widetilde{E}\setminus
br_{M({\rm k_{mc}})}$.  Then for all $E\subseteq R\subseteq
\overline{br_{M({\rm k_{mc}})}}$, $M[br_{M({\rm k_{mc}})}\cup
R]\models\bot$.

Suppose that there exists $E'\subseteq E$ with $x\not\in E'$ and
$E'\in E^{+}_{c}$. Then $E'\subset E$, and $br_{M({\rm k_{mc}})}\cup
E'\subseteq\overline{D}$, then  $M[br_{M({\rm k_{mc}})}\cup
E']\not\models\bot$. So, $E'\not\in E^{+}_{c}$.

Then we prove that $\bigcup E^{+}_{c}((M,\leq_{s}))\subseteq\bigcup
D^{-}_{c}((M,\leq_{s}))$ holds.  Let $E\in E^{+}_{c}((M,\leq_{s}))$,
then $E\cap br_{M({\rm k_{mc}})}=\emptyset$. We show that there
exists $D\in D^{-}_{c}$ with $x\in D$, for $x\in E$.

Consider $S=\{R\setminus\{x\}|E\subseteq
R\subseteq\overline{br_{M({\rm k_{mc}})}}\}$. Let $S'=\{T\in S|
M[br_{M_{\rm k_{mc}}}\cup T]\not\models\bot \}$.  Assume that
$S'=\emptyset$, then $M[br_{M_{\rm k_{mc}}}\cup
E\setminus\{x\}]\models\bot$, which contradicts $E\in
E^{+}_{c}((M,\leq_{s}))$.  So, $S'\not=\emptyset$

Let $T_{1}$ be a $\subseteq$-minimal set in $S'$ s.t. $M[br_{M_{\rm
k_{mc}}}\cup T_{1}]\not\models\bot$. Then $\overline{br_{M({\rm
k_{mc}})}\cup T_{1}}\in D^{-}_{c}(M)$, since for all $br_{M({\rm
k_{mc}})}\cup T_{1}\subset R$, $M[R]\models\bot$.

Furthermore, $x\not\in br_{M({\rm k_{mc}})}\cup T_{1}$, then $x\in
\overline{br_{M({\rm k_{mc}})}\cup T_{1}}$. $\ \ \ \ \ \ \ \ \ \ \ \
\ \ \ \ \ \  \Box$

This weak duality shows that if we consider bridge rules in the
maximal consistent section as reliable ones,  then diagnoses and
inconsistency explanations compatible with maximal consistent
section represent dual aspects of inconsistency caused by bridge
rules out of the maximal consistent section.

\section{Computational Complexity}

In this section we are concerned with the complexity aspects of
preferential multi-context systems.  We assume that the reader is
familiar with the classes $P$,  $NP$, and $coNP$ as well as
polynomial time hierarchy
($\Delta_{0}^{p}=\Sigma_{0}^{p}=\Pi_{0}^{p}=P$; and for all $k\geq
0$, $\Delta_{i+1}^{p}=P^{\Sigma_{i}^{p}},
\Sigma_{i+1}^{p}=NP^{\Sigma_{i}^{p}}, \Pi_{i+1}^{p}=
coNP^{\Sigma_{i}^{p}}$ )~\cite{Papadimitriou1994}.   We now
introduce the following classes:
\begin{itemize}
\item $D^{p}_{i}=\langle\Sigma_{i}^{p},\Pi_{i}^{p}\rangle$ is the class of all languages such that $L=L_{1}\cap
L_{2}$, where $L_{1}$ is in $\Sigma_{i}^{p}$ and  $L_{2}$ is in
$\Pi_{i}^{p}$ for all $i\geq 1$. In particular, $D^{p}_{1}$ is the
class of all languages such that $L=L_{1}\cap L_{2}$, where $L_{1}$
is in $NP$ and  $L_{2}$ is in $coNP$. The well known problem of
SAT-UNSAT is one of the canonical $D^{p}_{1}$-complete problems.
\item More generally, let $X_{1}$ and $X_{2}$ be two  complexity classes, then $\langle
X_{1},X_{2}\rangle$ is the class of all languages such that
$L=L_{1}\cap L_{2}$, where $L_{1}$ is in $X_{1}$ and $L_{2}$ is in
$X_{2}$.
\item Let $X$ be a complexity class,  $P^{X}$ (resp. $NP^{X}$) is the class of all languages that can be
recognized in polynomial time by a (resp. nondeterministic) Turing
machine equipped with an $X$ oracle, where an $X$ oracle solves
whatever instance of a problem in $X$ class in unit time.  In
particular, $P^{D_{k}^{p}[\log n]}$ is the class of all languages
can be recognized in polynomial time by a Turing machine using a
number of $D_{k}^{p}$ oracles bounded by a logarithmic function of
the size of input data.
\item $FX$ is the corresponding class of functions of $X$.
\end{itemize}

At first, we recall the complexity of calculating equilibria by
guessing so-called kernels of context belief sets presented
in~\cite{aaaiBrewkaE07}, and then we discuss the computational
complexity for calculating $l_{\leq}$-equilibria for preferential
multi-context systems based on that complexity result. Following
this, we discuss complexity aspects for identifying diagnoses and
inconsistency explanations compatible with the maximal consistent
section.

\subsection{Complexity for equilibria}

We consider the following aspects of computational complexity about
finding equilibria for preferential multi-context systems:
\begin{itemize}
\item {\em Consistency checking:}  the problem of deciding whether an PMCS $(M,\leq_{s})$ has an equilibrium.
\item {\em $l$-consistency checking:}  the problem of deciding whether an PMCS $(M,\leq_{s})=\langle M_{1},M_{2},\cdots, M_{m}\rangle$ has
an $l_{\leq}$-equilibrium for a given $1\leq l\leq m$.
\item {\em maximal $l$-consistency}: the problem of deciding whether  an PMCS $(M,\leq_{s})=\langle M_{1},M_{2},\cdots, M_{m}\rangle$ has a
maximal $l_{<}$-equilibrium  for a given $l$.
\item {\em maximal $l_{<}$-equilibrium}: the problem of computing  $l$ for  an PMCS $(M,\leq_{s})=\langle M_{1},M_{2},\cdots, M_{m}\rangle$ such
that it has a maximal $l_{<}$-equilibrium.
\end{itemize}

Note that the core of these  problems is to check consistency of
some sections or a whole preferential multi-context systems in
essences. However, the complexity aspects of calculating equilibria
by guessing so-called kernels of context belief sets has been
investigated in ~\cite{aaaiBrewkaE07}. In this paper, we also adopt
the following assumption of poly-size kernels about logics used in
multi-context systems presented in~\cite{aaaiBrewkaE07}.  A logic
$L$ has poly-size kernels, if there is a mapping $\kappa$ which
assigns to every $kb\in KB$ and $S\in ACC(KB)$ a set
$\kappa(kb,S)\subseteq S$ of size (written as a string) polynomial
in the size of $kb$, called the kernel of $S$, such that there is a
one-to-one correspondence $f$ between the belief sets in $ACC(kb)$
and their kernels, i.e., $S\rightleftharpoons
f(\kappa(kb,S))$~\cite{aaaiBrewkaE07}. Moreover, $L$ has kernel
reasoning in $\Delta_{k}^{p}$ if given any knowledge base $kb$, an
element $b$, and a set of elements $K$, deciding whether (i)
$K=\kappa(kb, S)$ for some $S\in ACC(kb)$ and (ii) $b\in S$ is in
$\Delta_{k}^{p}$~\cite{aaaiBrewkaE07}.

 Brewka
et al have pointed out that standard propositional non-monotonic
logics such as  DL, AEL, and NLP have poly-size kernels, moreover,
the standard propositional non-monotonic reasoning formalisms DL and
AEL have kernel reasoning in $\Delta_{2}^{p}$ ~\cite{aaaiBrewkaE07}.

Furthermore, for convenience, we assume that any belief set $S$ in
any logic $L$ contains a distinguished element ${\bf true}$; then
for $b={\bf true}$, (i) and (ii) together are equivalent to (i),
i.e., whether $K$ is a kernel for some acceptable belief set of
$kb$~\cite{aaaiBrewkaE07}.

Now we introduce the following theorem about computational
complexity about consistency checking for multi-context systems
based on assumptions above presented in~\cite{aaaiBrewkaE07}.

\begin{theorem}~\label{MCScomplexity}{\em ~\cite{aaaiBrewkaE07}} Given a finite MCS $M=(C_{1},C_{2},\cdots, C_{n})$
where all logics $L_{i}$ have poly-size kernels and kernel reasoning
in $\Delta_{k}^{p}$, deciding whether $M$ has an equilibrium is in
$\Sigma_{k+1}^{p}$.
\end{theorem}

Then we can get the following corollary about consistency checking
and proposition about $l$-consistency checking for preferential
multi-context systems directly from the theorem above, respectively.

\begin{corollary}~\label{SMCScomplexity}
Given a finite PMCS  $(M,\leq_{s})=\langle M_{1},M_{2},\cdots,
M_{m}\rangle$ where all logics $L_{i}$ have poly-size kernels and
kernel reasoning in $\Delta_{k}^{p}$, deciding whether
$(M,\leq_{s})$ has an equilibrium is in $\Sigma_{k+1}^{p}$.
\end{corollary}

\paragraph{Proof} Note that $(M,\leq_{s})$ has an equilibrium if and
only if $M$ has an equilibrium.  According to
Theorem~\ref{MCScomplexity}, the problem of deciding whether
$(M,\leq_{s})$ has an equilibrium is in $\Sigma_{k+1}^{p}$.  $\ \ \
\ \ \ \ \ \ \ \ \ \ \ \ \ \ \ \ \ \ \ \ \ \ \ \ \ \ \ \ \ \ \ \ \ \
\ \ \ \ \ \ \ \ \ \ \ \ \ \ \ \ \ \ \ \ \ \ \Box$

\begin{proposition}
Given a finite PMCS $(M,\leq_{s})=\langle M_{1},M_{2},\cdots,
M_{m}\rangle$ where all logics $L_{i}$ have poly-size kernels and
kernel reasoning in $\Delta_{k}^{p}$, deciding whether
$(M,\leq_{s})$ has an $l_{\leq}$-equilibrium for a given $1\leq
l\leq m$ is in $\Sigma_{k+1}^{p}$.
\end{proposition}

\paragraph{Proof} Given a number $1\leq l\leq m$,  $(M,\leq_{s})$ has an $l_{\leq}$-equilibrium if and
only if $M(l)$ has an equilibrium.  According to
Theorem~\ref{MCScomplexity}, the problem of deciding whether
$(M,\leq_{s})$ has an $l_{\leq}$-equilibrium is in
$\Sigma_{k+1}^{p}$. $\ \ \ \ \ \ \ \ \ \ \ \ \ \ \ \ \ \ \ \ \ \ \ \
\ \ \ \ \ \ \ \ \ \ \Box$

Next we give  the complexity of maximal $l$-consistency problem for
preferential multi-context systems in the case of given a positive
number $l$.

\begin{proposition}~\label{maximal-k-complexity1}
Given a finite PMCS $(M,\leq_{s})=\langle M_{1},M_{2},\cdots,
M_{m}\rangle$ where all logics $L_{i}$ have poly-size kernels and
kernel reasoning in $\Delta_{k}^{p}$, the problem of deciding
whether $(M,\leq_{s})$ has a maximal $l_{<}$-equilibrium for a given
$l<m$ is in ${D_{k+1}^{p}}$.
\end{proposition}

\paragraph{Proof} Note that $(M,\leq_{s})$ has a maximal $l_{<}$-equilibrium
for a given $l<m$ if and only if $M(l)$ has at least one equilibrium
and $M(l+1)$ has no equilibrium.  Recall the problem of deciding
whether $M$ has an equilibrium is in $\Sigma_{k+1}^{p}$, so, the
problem of deciding whether $M$ has no equilibrium is in
$\Pi_{k+1}^{p}$. Then
 the problem of deciding whether
$(M,\leq_{s})$ has a maximal $l_{<}$-equilibrium for a given $l<m$
is in $\langle\Sigma_{k+1}^{p},\Pi_{k+1}^{p}\rangle$, i.e.,
${D_{k+1}^{p}}$.   $\ \ \ \ \ \ \ \ \ \ \ \ \Box$

Then we are ready to get the following computational complexity for
the problem of maximal $l_{<}$-equilibrium.

\begin{proposition}~\label{computmaximalk} Given a finite PMCS $(M,\leq_{s})$ where all logics $L_{i}$ have poly-size kernels and kernel
reasoning in $\Delta_{k}^{p}$, the problem of computing  $l$ for
$(M,\leq_{s})$ such that it has maximal $l_{<}$-equilibrium is in
$FP^{D_{k+1}^{p}[\log n]}$.
\end{proposition}

\paragraph{proof} Consider the following algorithm for computing
$l$:
\begin{enumerate}
\item [(1)] if $M\not\models\bot$, then $l=m$;
\item [(2)] else if $M(1)\models\bot$, then $l=0$;
\item [(3)] else for $i$ from $1$ to $n-1$, if $(M,\leq_{s})$ has an
$i_{<}$-equilibrium, then $l=i$, break; else $i=i+1$.
\end{enumerate}

According to Corollary ~\ref{SMCScomplexity}, the problem of
deciding whether $M\not\models\bot$ is in $\Sigma_{k+1}^{p}$, and
checking whether $M(1)\models\bot$ is in $\Pi_{k+1}^{p}$.

From Proposition ~\ref{maximal-k-complexity1}, we have obtained that
the problem of deciding whether $(M,\leq_{s})$ has a maximal
$i_{<}$-equilibrium for each $i$ is in $D_{k+1}^{p}$. Therefore, $l$
can be computed in polynomial time by a Turing machine  equipped
with an $D^{p}_{k+1}$ oracle.  So, the problem of computing $l$ is
in $FP^{D_{k+1}^{p}}$.

Further, consider two particular singleton multi-contexts systems
$M_{1}=(C_{1})$ and $M_{2}=(C_{2})$, where both $L_{1}$ and $L_{2}$
are propositional logics with  $kb_{1}=kb_{2}=\{\}$,
$br_{1}=\{(1:a)\leftarrow \ {\bf not} \ (1:a)\}$ and
$br_{2}=\{(2:a)\leftarrow\}$.  Then
\begin{itemize}
\item $l=m$ if and only if  $\langle
M\not\models\bot,M_{1}\models\bot\rangle$ holds.
\item $l=0$ if and only if  $\langle
M_{2}\not\models\bot,M(1)\models\bot\rangle$ holds.
\end{itemize}
Moreover,  we can use a binary search on $\{1,2,\cdots, m-1\}$ to
find $l$ at step (3). Under such a case, $l$ can be computed by
using $O(\log_{2}m)$ calls to an $D_{p}^{k+1}$ oracle.  So, the
problem of computing $l$ is also in $FP^{D_{k+1}^{p}[\log n]}$.  $\
\ \ \  \ \ \ \ \ \ \  \ \ \Box$

Note that $M_{1\rightarrow l}$ is the maximal consistent section if
and only if $(M,\leq_{s})$ has a maximal $l_{<}$-equilibrium. Then
from  Proposition~\ref{computmaximalk}, we can get the following
complexity for the problem of identifying the maximal consistent
section of $(M,\leq_{s})$ directly.

\begin{corollary}~\label{maximalconset} Given a finite PMCS $(M,\leq_{s})$ where all logics $L_{i}$ have poly-size kernels and kernel
reasoning in $\Delta_{k}^{p}$, the problem of identifying  the
maximal consistent section of $(M,\leq_{s})$ is in
$FP^{D_{k+1}^{p}[\log n]}$.
\end{corollary}

We summarize these complexity aspects in Table
~\ref{complexityequilibrium}.

\begin{table}[h]
\caption{Complexity aspects about
equilibrium}~\label{complexityequilibrium}
\begin{center}
\begin{tabular}{l|c}
\hline
 {\small Problem} & {\small Complexity}\\
 \hline
{\small {\em consistency}} & $\Sigma_{k+1}^{p}$\\
 \hline
{\small {\em $l$-consistency}}  & $\Sigma_{k+1}^{p}$\\
\hline
{\small {\em maximal $l$-consistency}} & $D_{k+1}^{p}$   \\
\hline
 {\small {\em maximal $l_{<}$-equilibrium}} &  $FP^{D_{p}^{k+1}[\log
 n]}$\\
 \hline
\end{tabular}
\end{center}
\end{table}

\subsection{Computational complexity for diagnoses and explanations}

We focus on diagnoses and inconsistency explanations compatible with
the maximal consistent section in a preferential multi-context
system, respectively.

At first, we consider the following complexity aspects about
diagnoses and inconsistency explanations in the case that the
maximal consistent section is given.  Note that the complexity
aspects about finding diagnoses and inconsistency explanations for
multi-context systems have been investigated
in~\cite{DBLP:conf/kr/EiterFSW10}, respectively. The following
proposition shows that problems of finding diagnoses (resp.
inconsistency explanations) compatible with the  maximal consistent
section have the same complexity with that of finding diagnoses
(resp. inconsistency explanations) when the maximal consistent
section is given.

\begin{proposition}  Given a finite PMCS
$(M,\leq_{s})$ and its maximal consistent section $M_{1\rightarrow
{\rm k_{mc}}}$, deciding whether $D\subseteq br_{M}$ is a diagnosis
compatible with $M_{1\rightarrow {\rm k_{mc}}}$ has the same
computational complexity as consistency checking of $(M,\leq_{s})$.
\end{proposition}

\paragraph{Proof} Note that we only need to check whether  $M[br_{M}\setminus
D]\not\models\bot$ and $D\cap br_{M({\rm k_{mc}})}=\emptyset$.  $\ \
\ \ \ \ \ \ \ \ \ \ \ \ \ \ \ \ \ \ \ \ \ \ \ \ \ \ \ \ \ \ \ \ \ \
\ \ \ \ \ \ \ \ \ \ \ \ \ \ \ \ \ \ \ \ \ \ \ \ \ \ \ \ \ \ \Box$

\begin{proposition}  Given a finite PMCS
$(M,\leq_{s})$ and its maximal consistent section $M_{1\rightarrow
{\rm k_{mc}}}$, deciding whether $D\subseteq br_{M}$ is an
$c$-diagnosis has the same computational complexity as minimal
diagnosis recognition of $M$.
\end{proposition}

\paragraph{Proof} Note that the problem of deciding whether $D$ is
an $c$-diagnosis of $(M,\leq_{s})$ is equivalent to deciding whether
$D$ is a minimal $s$-diagnosis of $(M,\leq_{s})$ and $D\cap
br_{M({\rm k_{mc}})}=\emptyset$.     $\ \ \ \ \ \ \ \ \ \ \ \ \ \ \
\ \ \ \ \ \ \ \ \ \ \ \ \ \ \ \ \ \ \ \ \ \ \ \ \Box$

\begin{proposition}  Given a finite PMCS
$(M,\leq_{s})$ and its maximal consistent section $M_{1\rightarrow
{\rm k_{mc}}}$, deciding whether $E\subseteq br_{M}$ is an
$c$-inconsistency explanation  has the same computational complexity
as inconsistency explanation recognition of $M$.
\end{proposition}

\paragraph{Proof} Note that the problem of deciding whether $E$ is
an $c$-inconsistency explanation of $(M,\leq_{s})$ is equivalent to
deciding whether $E$ is an inconsistency explanation of
$(M,\leq_{s})$ and $E\cap br_{M({\rm k_{mc}})}=\emptyset$.     $\ \
\ \  \ \ \ \ \ \ \ \ \ \ \ \  \ \Box$

Now we consider the general case of finding $c$-diagnoses and
$c$-inconsistency explanations for preferential multi-context
systems.
 Let ${\rm C_{ms}}(M)$ be
the complexity for identifying  the maximal consistent section of
$(M,\leq_{s})$. For example, ${\rm C_{ms}}(M)$ is
$P^{D_{k+1}^{p}[\log n]}$ in the case illustrated in
Proposition~\ref{computmaximalk}. Let  ${\rm C_{sd}}(M,D)$ be the
complexity for deciding whether $D\subseteq br_{M}$ is an
$s$-diagnosis of $M$. Let ${\rm C_{e}}(M,E)$ is the complexity for
deciding whether $E$ is an inconsistency explanation of $M$. We
assume that ${\rm C_{e}}(M,E)$ is closed under conjunction,
according to discussion about such complexity aspects
in~\cite{DBLP:conf/kr/EiterFSW10}.

\begin{proposition}Given a finite PMCS
$(M,\leq_{s})$, deciding whether $D\subseteq br_{M}$ is an
$c$-diagnosis is in $\langle {\rm C_{ms}}(M),{\rm C_{sd}}(M,D)
\rangle$.
\end{proposition}

\paragraph{Proof} To decide whether $D\subseteq br_{M}$ is an
$c$-diagnosis, we need to check
\begin{enumerate}
\item [(1)] whether $D\cap br_{M({\rm k_{mc}})}=\emptyset$ holds;
\item [(2)]whether $D$
is an $s$-diagnosis.
\end{enumerate}

Note that   $D\cap br_{M({\rm k_{mc}})}=\emptyset$ if and only if
$br_{M({\rm k_{mc}})}\subseteq\overline{D}$. Then (1) is equal to
finding the maximal consistent section from sections not involved in
$D$. So, the problem of deciding whether $D\subseteq br_{M}$ is an
$c$-diagnosis is in $\langle {\rm C_{ms}}(M),{\rm C_{sd}}(M,D)
\rangle$. $\ \ \Box$

\begin{proposition}~\label{c-incon-explan-compl}Given a finite PMCS
$(M,\leq_{s})$, deciding whether $E\subseteq br_{M}$ is an
$c$-inconsistency explanation is in $\langle {\rm C_{ms}}(M),{\rm
C_{e}}(M,E) \rangle$, where ${\rm C_{e}}(M,E)$ is the complexity for
deciding whether $E$ is an inconsistency explanation of $M$.
\end{proposition}

\paragraph{Proof} To decide whether $E\subseteq br_{M}$ is an
$c$-inconsistency explanation, we need to check
\begin{enumerate}
\item [(1)] whether $E\cap br_{M({\rm k_{mc}})}=\emptyset$ holds;
\item [(2)]whether $E$
is an inconsistency explanation.
\end{enumerate}
Note that    (1) is equal to finding the maximal consistent section
from sections not involved in $E$. So, the problem of deciding
whether $E\subseteq br_{M}$ is an $c$-inconsistency explanation is
in  $\langle {\rm C_{ms}}(M),{\rm C_{e}}(M,E) \rangle$. $\ \ \ \ \ \
\Box$

\begin{proposition}Given a finite PMCS
$(M,\leq_{s})$, deciding whether $E\subseteq br_{M}$ is in
$E^{+}_{c}((M,\leq_{s}))$
 is in $\langle\langle {\rm C_{ms}}(M),{\rm C_{e}}(M,E)
\rangle,{\rm coC_{e}}(M,E)\rangle$.
\end{proposition}

\paragraph{Proof} Note that $E\in E^{+}_{c}((M,\leq_{s}))$ if and
only if $E$ is an $c$-inconsistency explanation and $E$ is minimal
w.r.t. $\subseteq$. We have obtained that deciding whether
$E\subseteq br_{M}$ is an $c$-inconsistency explanation is in
$\langle {\rm C_{ms}}(M),{\rm C_{e}}(M,E) \rangle$ in Proposition
~\ref{c-incon-explan-compl}.

From Lemma 2 in~\cite{DBLP:conf/kr/EiterFSW10}, we can check
subset-minimality of $E$ by deciding whether none of
$E\setminus\{x\}$ is an inconsistency explanation for all $x\in E$.
Note that the number of these checks is linear w.r.t. $|E|$, and
${\rm C_{e}}(M,E)$ is closed under conjunction.  So, deciding
whether $E\in E^{+}_{c}((M,\leq_{s}))$
 is in $\langle\langle {\rm C_{ms}}(M),{\rm C_{e}}(M,E)
\rangle,{\rm coC_{e}}(M,E)\rangle$. $\ \ \ \ \ \ \ \ \ \ \ \ \ \ \
\Box$

Note that we consider the general case of complexity. It is not
difficult to consider the usual cases discussed
in~\cite{DBLP:conf/kr/EiterFSW10}.

\section{Comparison and Discussion}

Preferential multi-context systems provide a framework for
incorporating preferences on contexts in multi-context systems.
 However, the following aspects distinguish preferential
multi-context systems from the original multi-context systems
presented in~\cite{aaaiBrewkaE07}. At first, the compatibility of a
total preorder relation over contexts  with a multi-context system
imposes a constraint on bridge rules, i.e., any appearance of less
preferred contexts is prohibited in the body of a bridge rule for a
given context. Only one-way information flow between any two strata
is allowed to occur in a preferential multi-context system. The
intuition behind this constraint is that less reliable information
cannot be used to revise more reliable knowledge.
 Second, one-way information flow
makes any section of a preferential multi-context system capable of
capturing all the  information exchange among contexts in that
section. This signifies that each section is also a meaningful
preferential multi-context system.  Third, preferential
multi-context systems are concerned with partial equilibria such as
$l_{\leq}$-equilibria as well as equilibria.

Note that preferential multi-context systems also analyze
inconsistency in terms of diagnoses and inconsistency explanations
presented in ~\cite{DBLP:conf/kr/EiterFSW10}. However, allowing for
the role of preferences on  contexts in a given preferential
 multi-context system, we are more interested in diagnoses and
inconsistency explanations compatible with the maximal consistent
section. More interestingly, diagnoses and inconsistency
explanations compatible with the maximal consistent section have
duality relation.  This implies that the compatibility with the
maximal consistent section does not destroy
 the duality relation between diagnoses and inconsistency
explanations~\cite{DBLP:conf/kr/EiterFSW10}.

Actually, the compatibility of diagnoses (resp. inconsistency
explanations) with the maximal consistent section essentially
provides a way to discriminate between all diagnoses (resp.
inconsistency explanation) based on preferences on contexts. In this
sense, such a compatibility can be considered as some kind of filter
to filter some undesirable diagnoses (resp. inconsistency
explanations) ~\cite{jelia/EiterFW10}.

Preferential multi-context systems aim to address the total preorder
relation over contexts.  However, combining preferences with
contexts is one of the important issues in integrating and sharing
contextual knowledge~\cite{iclpBrewka07}.  Moreover, as stated
in~\cite{aim/BrewkaNT08}, there is a multifaceted relationship
between nonmonotonic logics and preferences.  As a framework for
integrating arbitrary monotonic and nonmonotonic logics, it is
necessary to incorporating such a relation in multi-context systems.
This may be one of directions of our  future work.

\section{Conclusion}

In this paper, we have presented the preferential multi-context
system, which provides a promising framework for combining
multi-context systems with the total preorder relations on their
contexts. Preferential multi-context systems take into account  the
impact of preferences among contexts on their inter-contextual
information exchange.  Only information flow from more preferred
contexts to less preferred ones is allowed to occur in preferential
multi-context systems.  In such a preferential multi-context system,
a context may be revised based on only information exchange with
more or equally preferred contexts.

This paper presented the following contributions to multi-context
systems community:
\begin{itemize}
\item We proposed the notion of preferential multi-context system,
which consists of a multi-context system with a total preorder
relation compatible with that system.
\item We extended the equilibrium semantics for multi-context systems and proposed a notion of $l_{\leq}$-equilibrium representing belief
states acceptable for at least contexts of the first $l$ strata in a
preferential multi-context system. Furthermore, we proposed a notion
of maximal $l_{<}$-equilibrium describing belief states acceptable
for contexts in the maximal consistent section of a preferential
multi-context system.
\item We proposed inconsistency diagnoses and inconsistency
explanations compatible with the maximal consistent section,
respectively. Moreover, we discussed their duality relation.
\item We investigated the computational complexity aspects for
calculating $l_{\leq}$-equilibria and identifying diagnoses and
inconsistency explanations compatible with the maximal consistent
section, respectively.
\end{itemize}

%% The Appendices part is started with the command \appendix;
%% appendix sections are then done as normal sections
%% \appendix

%% \section{}
%% \label{}

%% References
%%
%% Following citation commands can be used in the body text:
%% Usage of \cite is as follows:
%%   \cite{key}          ==>>  [#]
%%   \cite[chap. 2]{key} ==>>  [#, chap. 2]
%%   \citet{key}         ==>>  Author [#]

%% References with bibTeX database:

\bibliographystyle{plain}
\bibliography{SMCS2013}

%% Authors are advised to submit their bibtex database files. They are
%% requested to list a bibtex style file in the manuscript if they do
%% not want to use model1a-num-names.bst.

%% References without bibTeX database:

% \begin{thebibliography}{00}

%% \bibitem must have the following form:
%%   \bibitem{key}...
%%

% \bibitem{}

% \end{thebibliography}

\end{document}